\relax
\documentclass[letterpaper]{article} 
\usepackage{aaai22}  
\usepackage{times}  
\usepackage{helvet}  
\usepackage{courier}  
\usepackage[hyphens]{url}  
\usepackage{graphicx} 
\usepackage{natbib}  
\usepackage{caption} 
\DeclareCaptionStyle{ruled}{labelfont=normalfont,labelsep=colon,strut=off} 
\usepackage{algorithm}
\usepackage{algorithmic}

\usepackage{newfloat}
\usepackage{listings}
\floatstyle{ruled}
\newfloat{listing}{tb}{lst}{}
\floatname{listing}{Listing}
\nocopyright
\usepackage{booktabs}
\usepackage{amsfonts,amssymb}
\usepackage{multirow}
\usepackage{xspace}
\usepackage{subfigure}
\usepackage{color}
\usepackage{amsmath}
\usepackage{stmaryrd}
\usepackage{stfloats}
\usepackage{makecell}
\usepackage{graphicx}

\newcommand{\ie}{\emph{i.e.,}\xspace}
\newcommand{\eg}{\emph{e.g.,}\xspace}

\title{Go Wider Instead of Deeper}

\author{
    Fuzhao Xue, 
        Ziji Shi,
        Futao Wei,
        Yuxuan Lou, 
        Yong Liu, 
        Yang You \\
}
\affiliations{
    Department of Computer Science, National University of Singapore, Singapore \\
    \{f.xue,ziji.shi\}@u.nus.edu, weifutao2019@gmail.com, yuxuanlou@u.nus.edu,  \{liuyong,youy\}@comp.nus.edu.sg
}

\usepackage{bibentry}

\begin{document}

\maketitle

\begin{abstract}
More transformer blocks with residual connections have recently achieved impressive results on various tasks. To achieve better performance with fewer trainable parameters, recent methods are proposed to go shallower by parameter sharing or model compressing along with the depth. However, weak modeling capacity limits their performance. Contrastively, going wider by inducing more trainable matrixes and parameters would produce a huge model requiring advanced parallelism to train and inference.

In this paper, we propose a parameter-efficient framework, going wider instead of deeper. Specially, following existing works, we adapt parameter sharing to compress along depth. But, such deployment would limit the performance. To maximize modeling capacity, we scale along model width by replacing feed-forward network (FFN) with mixture-of-experts (MoE). Across transformer blocks, instead of sharing normalization layers, we propose to use individual layernorms to transform various semantic representations in a more parameter-efficient way. To evaluate our plug-and-run framework, we design WideNet and conduct comprehensive experiments on popular computer vision and natural language processing benchmarks. On ImageNet-1K, our best model outperforms Vision Transformer (ViT) by $1.5\%$ with $0.72 \times$ trainable parameters. Using $0.46 \times$ and $0.13 \times$  parameters, our WideNet can still surpass ViT and ViT-MoE by $0.8\%$ and $2.1\%$, respectively. On four natural language processing datasets, WideNet outperforms ALBERT by $1.8\%$ on average and surpass BERT using factorized embedding parameterization by $0.8\%$ with fewer parameters.\footnote{We will release our code upon acceptance.}

\end{abstract}

\section{Introduction}
\label{sec:intro}

\noindent Transformer-based models have achieved promising results on various tasks (\eg Q\&A~\cite{qu2019bert,yang2020bert}, relation extraction~\cite{xue2020gdpnet,xue2020embarrassingly,zhou2020document}). To further improve the effectiveness and efficiency of the transformer, there are two trains of thought to deploy trainable parameters. The first thought is to scale transformer along width to more trainable parameters (\eg Switch Transformer \cite{fedus2021switch}, ViT-MoE~\cite{riquelme2021scaling}). These sparse models can scale to extremely large models with comparable FLOPs by sparse conditional computation. Another thought is to decrease the trainable parameters for a lite model. To this end, some works propose to reuse the trainable parameters across transformer blocks (\eg Universal Transformer~\cite{dehghani2018universal} and ALBERT~\cite{lan2019albert}). Model compression~\cite{xu-etal-2020-bert,sun-etal-2019-patient} can also make transformer more parameter efficient.


The two existing methods both have their own limitations. For huge models, one typical and effective method to scale trainable parameters is replacing part of the feed-forward network (FFN) layer in transformer blocks with MoE layers. In each MoE layer, to refine one single token representation, only a few experts are activated, so the MoE based transformer holds comparable FLOPs with the vanilla transformer. However, during training and inference, we are required to use advanced parallelisms (\eg tensor~\cite{shoeybi2019megatron}, sequence~\cite{li2021sequence}, pipeline~\cite{huang2018gpipe} and expert parallelism~\cite{lepikhin2020gshard}) to hold these models on TPU or GPU. Also, the performance cannot improve linearly during scaling. Another limitation is that the sparseness of MoE based models cannot scale well on relatively small datasets. We will discuss the reason for this phenomenon in the following sections. For small models, although they can reduce trainable parameters significantly by going shallower, the performance of these shallower models is still under the original transformers. These smaller models are constructed by compressing the original model along with depth so all transformer blocks share the same knowledge. Such structure induces the unavoidable loss of model capacity.

In this paper, we present a parameter deployment framework that deploys trainable parameters more effectively: going wider instead of deeper. We then implement it on the transformer and named it as WideNet.  Specially, we first employs parameter sharing along with depth to go shallower. Due to avoidable model capacity loss, we go wider by using the same MoE layer in all transformer blocks. The multi-head attention layer is also shared across the blocks. To help the transformer blocks learn different semantics and maximize the modeling capacity from MoE layer, we do not share the normalization layers. Different trainable parameters of the normalization layer enable transformer blocks to be fed by diversified representations. Since the modeling capacity of each transformer block has been enhanced by the MoE layer, it can model diversified semantics effectively with the same trainable parameters. Therefore, with one attention layer and one single stronger MoE layer learning complex representations, and independent normalization layers for diversified semantic representations, going wider instead of deeper is a more parameter-efficient and effective framework.

Compared with simply scaling along the width, going wider instead of deeper is a more parameter-efficient framework, which makes the models small enough to be adapted to downstream tasks without advanced parallelisms. Second, each expert in WideNet can be trained by more token representations so that it has better generalization performance.

Compared with the models simply compressed along with the depth, all transformer blocks in WideNet share one same MoE layer instead of one FFN layer. Such structure maximizes the modeling ability of every transformer block. More experts can model more complex token representations with a stronger capacity. Another difference is the independent normalization layers. These layers come with few additional trainable parameters, but they can transform input representations to other semantic domains. In this case, with a strong enough single MoE layer, WideNet can still model semantics from different levels well. Moreover, in every transformer block, each expert only receives a part of token representations that usually correspond to different input tokens. 

\begin{figure*}[t]
\centering
\includegraphics[width=0.9\textwidth]{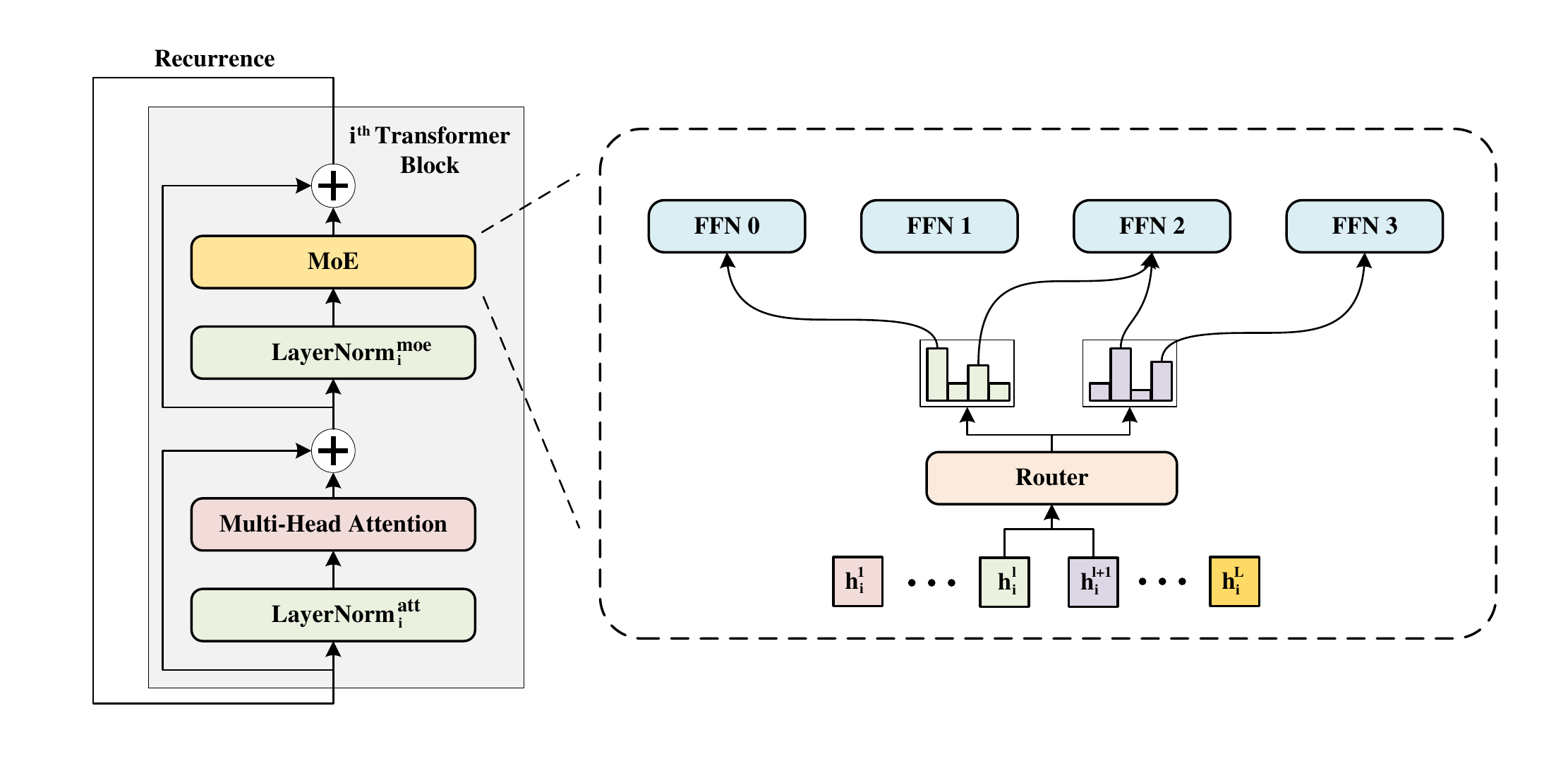}
\caption{The overall architecture of the proposed WideNet. Compared with vanilla transformer, we replace FFN layer by MoE layer and share the trainable parameters except the normalization layers.}
\label{fig:overview}
\end{figure*}

Our contributions are summarized as three folds:

\begin{itemize}

\item To improve the parameter efficiency, we propose sharing the MoE layer across transformer blocks. The shared experts can receive diversified token representations in different transformer blocks, which enables each expert to be fully trained.


\item We propose to keep individual normalization layer across transformer blocks. The individual normalization layers can transform input hidden vectors to semantic information by adding few trainable parameters. Then, diversified input can be fed into the same attention layer or stronger MoE layer to model different semantics.  


\item By combing the two thoughts above, we propose going wider instead of deeper, a more parameter-efficient and effective framework. We then implement this framework as WideNet and evaluate it on both computer vision and natural language processing tasks. Due to the more efficient parameter deployment, WideNet outperforms baselines with less trainable parameters. We expect our WideNet can serve as a next-generation transformer backbone.


\end{itemize}

\section{Mixture-of-Experts}
\label{sec:moe}

In this paper, we focus on a novel trainable parameter deployment framework and implement this framework on the transformer as WideNet. The overall structure is shown in Fig.~\ref{fig:overview}. We use Vision Transformer as the backbone in this example, which means we normalize the representations before the attention layer or FFN layer. We also extend WideNet to other transformer models (\eg BERT~\cite{devlin-etal-2019-bert}) in this paper. In WideNet, we replace the FFN layer with the MoE layer. Parameter sharing across transformer blocks is employed for a more parameter-efficient deployment. Within each MoE layer, we have one router to select $K$ experts to learn more complex representations. Please note the trainable parameters in layer normalization are not shared for more diversified semantic representations.




\subsection{Conditional Computation with MoE}

Our core idea is to deploy more trainable parameters along the width and fewer trainable parameters along with the depth. To this end, we employ MoE to scale transformer along with width. As a typical conditional computation model~\citep{bengio2013deep}, MoE only activates a few experts, \ie subsets of a network. For each input, we feed only a part of hidden representations required to be processed into the selected experts. 


Following \citet{shazeer2017outrageously}, given $E$ trainable experts and input representation $x\in \mathbb{R}^D$, the output of MoE model can be formulated as:

\begin{equation}
\mathrm{MoE}(x)=\sum_{i=1}^E {g(x)}_i {e(x)}_i
\end{equation}
where ${e(\cdot)}_i$ is a non-linear transformation $\mathbb{R}^D \to \mathbb{R}^D$ of $i^{\mathrm{th}}$ expert, and ${g(\cdot)}_i$ is $i^{\mathrm{th}}$ element of the output of trainable router $g(\cdot)$, a non-linear mapping $\mathbb{R}^D \to \mathbb{R}^E$. Usually, both $e(\cdot)$ and $g(\cdot)$ are parameterized by neural networks. 

According to the formulation above, when $g(\cdot)$ is a sparse vector, only part of experts would be activated and updated by back-propagation during training. In this paper, for both vanilla MoE and our WideNet, each expert is an FFN layer.


%

\subsection{Routing}

To ensure a sparse routing $g(\cdot)$, we use $\mathrm{TopK}()$ to select the top ranked experts. Then, following \citet{riquelme2021scaling}, $g(\cdot)$ can be written as:

\begin{equation}\label{eq:topK}
\mathrm{g}(x)=\mathrm{TopK}(\mathrm{softmax}(f(x)+\epsilon))
\end{equation}
where $f(\cdot)$ is routing linear transformation $\mathbb{R}^D \to \mathbb{R}^E$, and $\epsilon \sim \mathcal{N}(0,\frac{1}{E^2})$ is a Gaussian noise for exploration of expert routing. We use $\mathrm{softmax}$ after $f(\cdot)$ for better performance and more sparse experts \citep{riquelme2021scaling,fedus2021switch}. When $K \ll E$, most elements of $\mathrm{g}(x)$ would be zero so that sparse conditional computation is achieved.

\subsection{Balanced Loading}


In MoE based transformer, we dispatch each token to $K$ experts. During training, if the MoE model has no regularization, most tokens may be dispatched to a small portion of experts. Such an unbalanced assignment would decrease the throughput of the MoE model. In addition, more importantly, most additional trainable parameters would not be fully trained so that the sparse conditional model cannot surpass the corresponding dense model during scaling. Therefore, for balanced loading, we have two things to avoid: (1) too many tokens dispatched to one single expert, and (2) too few tokens received by one single expert. To solve the first issue, buffer capacity $B$ is required. That is, for each expert, we only preserve $B$ token at most regardless of how many tokens are dispatched to this expert. If more than $B=CKNL$ tokens are assigned, the left tokens would be dropped. $C$ is the capacity ratio, a pre-defined hyperparameter to control the ratio of tokens preserved for each expert. Usually, $C\in[1,2]$, and we set $C$ as 1.2 when no special explanation is used. $K$ is the number of selected experts for each token. $N$ is the batch size on each device\footnote{For easier using on downstream tasks, we implement our method with only data parallelism.}. $L$ is the sequence length. For computer vision tasks, $L$ denotes the number of patch tokens in each image. 

Buffer capacity $B$ helps us drop redundant tokens for each expert to maximize throughput but it cannot ensure all experts to receive enough token to train. In other words, until now, the routing is still unbalanced. Therefore, we follow \citet{fedus2021switch} to use a differentiable load balance loss instead of separate load-balancing and importance-weighting losses for a balanced loading in the router. For each routing operation, given $E$ experts and $N$ batches with $NL$ tokens, the following auxiliary loss is added to the total model loss during training: 


\begin{equation}\label{eq:balance_loss}
l_{balance} = E \cdot \sum_{i=1}^E m_i \cdot P_i
\end{equation}
where $m$ is vector. $i^{\mathrm{th}}$ element is the fraction of tokens dispatched to expert $i$:

\begin{equation}
m_i = \frac{1}{L} \sum_{j=1}^{L} \mathrm{h}(x_j)_i
\end{equation}
where $\mathrm{h}(\cdot)$ is a index vector selected by $\mathrm{TopK}$ in Eq.~\ref{eq:topK}. $\mathrm{h}(x_j)_i$ is $i^{\mathrm{th}}$ element of $\mathrm{h}(x_j)$. It is noticeable that, different from $g(x)_i$ in Eq.~\ref{eq:topK}, $m_i$ and $\mathrm{h}(x_j)_i$ are non-differentiable. However, a differentiable loss function is required to optimize MoE in an end-to-end fashion. Therefore, we define $P_i$ in Eq.~\ref{eq:balance_loss} as:

\begin{equation}
P_i = \mathrm{softmax}(f(x)+\epsilon)_i
\end{equation}

We can observe $P_i$ is $i^{\mathrm{th}}$ element of routing linear transformation after $\mathrm{softmax}$ activation function, and $P_i$ is differentiable. 

The goal of load balancing loss is to achieve a balanced assignment. When we minimize $l_{balance}$, we can see both $m$ and $P$ would close to a uniform distribution.

\section{Go wider instead of deeper}
\label{sec:wide}

\subsection{Sharing MoE across transformer blocks}

As shown in Fig.~\ref{fig:overview}, WideNet adopts parameter sharing across transformer blocks to improve parameter efficiency, and MoE layer is used to improve model capacity. In addition, as we use the MoE layer to obtain a stronger modeling ability, to overcome the overfitting from sparse conditional computation, we are required to feed enough tokens to each expert. To this end, WideNet uses the same router and experts in different transformer blocks. Formally, given hidden representations $H^1=\{h_1^1,h_2^1,\dots,h_L^1\}$ as input of the first transformer block, we can define the parameter sharing as $H^{i+1}=\mathrm{MoE}(H^i)$, which is different from the existing MoE based models $H^{i+1}=\mathrm{MoE}^i(H^i)$. Please note that, although we share trainable parameters in the MoE layer including the router, token representations corresponding to the same token are different in every transformer block. That is, $h_i^j$ and $h_i^{j+1}$ may be dispatched to different experts. Therefore, each expert would be trained by more varied tokens for better generalization performance.

\subsection{Individual Layer Normalization}
Although existing works~\citep{lan2019albert} show that the activations in different transformer blocks are similar, the cosine distance is still much larger than zero. Therefore, different from existing works~\citep{dehghani2018universal,lan2019albert} sharing all weights across transformer blocks, to encourage more diversified input representations of different blocks, we only share multi-head attention layer and FFN (or MoE) layer, which means trainable parameters of layer normalization are different across blocks. 

In summary, $i^{\mathrm{th}}$ transformer block in our framework can be written as:
\begin{equation}
\begin{aligned}
x' &= \mathrm{LayerNormal}^{att}_{i}(x) \\
x &= \mathrm{MHA}(x') + x\\
x''&= \mathrm{LayerNormal}^{moe}_{i}(x) \\
x &= \mathrm{MoE}(x'') + x\\
\end{aligned}
\end{equation}
The normalization layer $ \mathrm{LayerNormal}(\cdot)$ is:

\begin{equation}\label{eq:layer_norm}
\mathrm{LayerNormal}(x)=\frac{x-\mathrm{E}[x]}{\sqrt{\mathrm{Var}[x]+\epsilon}}*\gamma + \beta 
\end{equation}
where $\gamma \in \mathbb{R}^D$ and $\beta \in \mathbb{R}^D$ are two trainable vectors. Layer normalization only requires these two small vectors so individual normalization would just add few trainable parameters into our framework. We can find the difference between shared layer normalization and the individual ones is the mean and magnitude of output. For shared layer normalization, the input of MHA and MoE layer are more similar in different transformer blocks. Since we have shared trainable matrixes, we encourage more diversified input to represent various semantics in different transformer blocks.

\subsection{Optimization}

Although we reuse the trainable parameters of the router in every transformer block, the assignment would be different due to different input representations. Therefore, given $T$ times routing operation with the same trainable parameters, we have the following loss for optimization:

\begin{equation}
loss=l_{main} + \lambda \sum_{t=1}^T l_{balance}^T
\end{equation}
where $\lambda$ is a hyper-parameter to ensure a balanced assignment, and we set it as a relatively large number, \ie 0.01 in this work. Similar to existing MoE based models, we found the performance is non-sensitive to $\lambda$. $l_{main}$ is the main target of our transformer. For example, on supervised image classification, $l_{main}$ is cross-entropy loss.


\section{Experiments}
\label{sec:experiments}




\subsection{Computer Vision}
\label{sec:cv}

\begin{table}[t]
\centering
\caption{Results on ImageNet-1K pretraining.}
\label{tbl-main-pretraining-ImageNet}
\begin{tabular}{l|l l l}
\toprule
Model     & Parameters & ImageNet-1K  \\ \midrule
ViT-B               & 87M       & 78.6        \\
ViT-L              & 305M         & 77.5   \\ \midrule
ViT-MoE-B           & 128M          & 77.9   \\
ViT-MoE-L           & 406M        & 77.4   \\ \midrule
WideNet-B         & 29M        & 77.5       \\
WideNet-L          & 40M     & 79.5       \\
WideNet-H          & 63M      & \textbf{80.1}      \\
\bottomrule
\end{tabular}
\end{table}

\subsubsection{Experimental Settings}
\label{sec:cv-settings}

We use ILSVRC-2012 ImageNet~\citep{deng2009imagenet} and Cifar10~\citep{krizhevsky2009learning} as platforms to evaluate our framework. ImageNet we used in this work has 1k classes and 1.3M images. We denote it as ImageNet-1K in the following experiments. We select ViT~\citep{dosovitskiy2020image} and ViT-MoE~\citep{riquelme2021scaling} as baselines. We first reimplement ViT by Tensorflow 2.x and tune it to a reasonable performance. For all models in this section, we use Inception-style pre-processing, Mixup~\citep{zhang2017mixup}, RandAugment~\citep{cubuk2020randaugment} and label smoothing~\citep{szegedy2016rethinking,yuan2020revisiting} as data augmentation. We also observe that AdamW optimizer~\citep{loshchilov2017decoupled} is sensitive to hyper-parameters and learning schedules. LAMB optimizer~\citep{you2019large} can achieve comparable performance but it is more robust to the hyper-parameters. For fair comparison, following \citet{zhai2021scaling}, we evaluate WideNet on three scales (\ie WideNet-Base, WideNet-Large and WideNet-Huge). The attention and FFN dimensions of different scales are the same as ViT-MoE except for WideNet-B. For WideNet-B, we use a hidden dimension of FFN as 4096 instead of 3072 for a more stable training.

Instead of achieving SoTA performance, the goal of this paper is to show that our parameter deployment framework can improve the transformer backbone with less trainable parameters. Therefore, we employ LAMB instead of AdamW for more general and typical experiments. For MoE based models (\ie ViT-MoE and WideNet), we set the weight of load balance loss $\lambda$ as 0.01. Without special instructions, we use 4 experts in total and Top 2 experts selected in each transformer block. The capacity ratio $C$ is set as 1.2 for a trade-off between accuracy and speed. We pretrain our models on 256 TPUv3 cores. According to recent work~\citep{zhai2021scaling}, different types of the prediction head have no significant difference on ImageNet's few-shot performance. We also verify this conclusion on training ImageNet from scratch. In this work, for ViT, we use the typical token head, which means we insert [CLS] token at the start of patch tokens and use it to classify the image. For MoE based models, to fully use the token representations after the final MoE layer, we employ a global average pooling head instead of the token head. 

During finetuning, we still follow \cite{dosovitskiy2020image} and use SGD optimizer with momentum. Compared with pretraining on ImageNet-1K, label smoothing and warm-up are removed.

\subsubsection{Comparison with baselines}

We follow the hyper-parameter setting of baselines in pretraining and finetuning for a fair comparison. Please see Appendix for details. Such implementation also shows that our model is robust to hyper-parameters.

We report the Top-1 accuracy on ImageNet-1K in Table ~\ref{tbl-main-pretraining-ImageNet} and Cifar10 in Appendix. Observe that WideNet-H achieves the best performance and significantly outperforms ViT and ViT-MoE models on ImageNet-1K. Compared with the strongest baseline, our WideNet-H outperforms ViT-B by $1.5\%$ with less trainable parameters. Even if we use the smallest model, WideNet-B, it still achieves comparable performance with ViT-L and ViT-MoE-B with over $4 \times$ less trainable parameters. When we scale up to WideNet-L, it has surpassed all baselines with half trainable parameters of ViT-B and $0.13 \times$ parameters of ViT-L. 

Another observation is, unlike training MoE based models on huge datasets (\eg JFT-300M~\citep{sun2017revisiting} and C4~\citep{raffel2019exploring}), MoE cannot benefit ViT on ImageNet-1K, which is 200 times smaller than original ViT-MoE used in pretraining\footnote{This dataset is not publicly available.}.

\subsection{Natural Language Processing}
\label{sec:nlp}

\begin{table*}[t]
\centering
\caption{Results of funetuning on GLUE benchmarks}
\label{tbl-nlp-finetune}
\begin{tabular}{l|lllll|l}
\toprule
Model                   & \#para & SQuAD1.1 & SQuAD2.0 & MNLI & SST-2  & Avg\\ \midrule
ALBERT                & 12M    & 89.3/82.3 & 80.0/77.1 & 81.5 & 90.3 & 84.0 \\
BERT                  & 89M    & 89.9/82.8 & 80.3/77.3 & 83.2 & 91.5  & 85.0\\  \midrule
WideNet 4 experts  & 26M    & 89.6/82.7 & 80.6/77.4 &  82.6 & 91.1  & 84.7 \\ 
WideNet 8 experts  & 45M    & 90.0/82.7 & 80.6/77.7 & 83.3 & 91.9 &  85.2 \\ 

WideNet 16 experts & 83M    & \textbf{90.9/83.8} & \textbf{81.0/77.9} & \textbf{84.1} & \textbf{92.2}  & \textbf{85.8}\\
\bottomrule
\end{tabular}
\end{table*}

The main contribution of this work is to design a more parameter-efficient and plug-in framework for various AI applications. Therefore, we further evaluate our work on natural language processing (NLP) after computer vision (CV). The training of experiments on NLP can still be splitted into 2 stages, pretraining and finetuning. 

\subsubsection{Experimental Settings}
\label{sec:nlp-settings}

Following BERT~\citep{devlin-etal-2019-bert} and ALBERT~\citep{lan2019albert}, in this section, we pretrain all models by English Wikipedia~\citep{devlin-etal-2019-bert} and BOOKCORPUS~\citep{zhu2015aligning}. Since the goal of this work is to design a parameter-efficient framework, all models including BERT use factorized embedding parameterization. That is, the WordPiece embedding size $E$ is 128. The hyperparameter settings of experiments on NLP can be found in Appendix, which is the same as ALBERT for a fair comparison. Similar to the experiments on vision tasks, we pretrain our models by LAMB on 256 TPUv3 cores. The learning rate is 0.00176, which is the same as ALBERT claimed~\citep{you2019reducing}.

During finetuning, we evaluate our model on the General Language Understanding Evaluation (GLUE) benchmark~\citep{wang-etal-2018-glue}, two versions of the Stanford Question Answering (SQuAD) dataset~\citep{rajpurkar-etal-2016-squad,rajpurkar-etal-2018-know}. For GLUE experiments, we report median over 5 runs on development set because of relatively large variance.

\subsubsection{Downstream Evaluation}
\label{sec:nlp-downsteam}

Different from the experiments on CV, we report the evaluation results on downstream tasks directly in this section. As shown in Table~\ref{tbl-nlp-finetune}, when we use more experts, our WideNet outperforms ALBERT by a large margin. For instance, WideNet with 4 experts surpasses ALBERT by $1.2\%$ in average. When we increase the number of experts $E$ to 16 to achieve slightly less trainiable parameters than BERT with factorized embedding  parameterization, our WideNet also outperforms it on all four downstream tasks, which shows the parameter-efficiency and effectiveness of going wider instead of deeper.


\begin{figure*}
\centering
\begin{minipage}[b]{.45\textwidth}
\includegraphics[width=\textwidth]{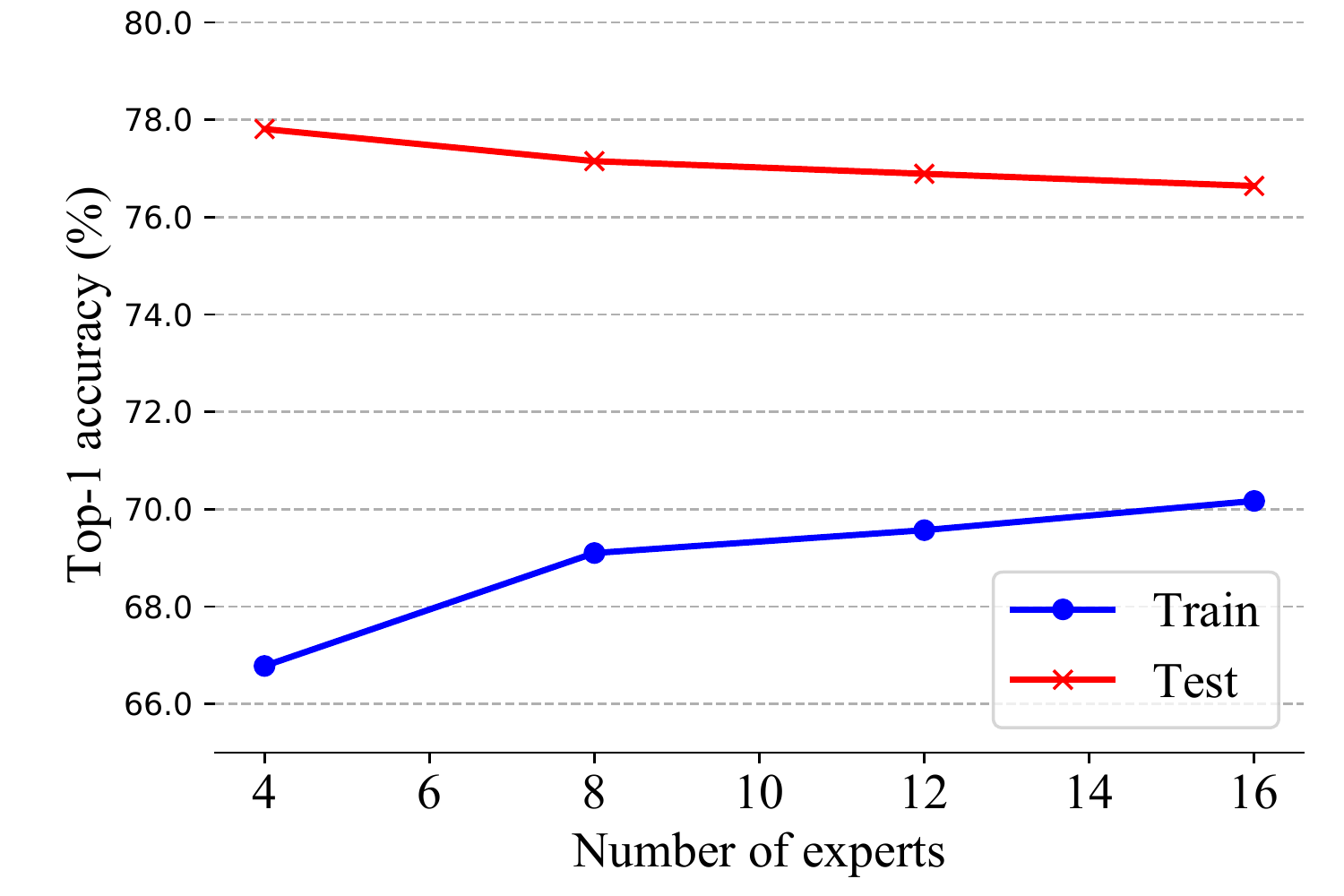}
\caption{Results of scaling the number of experts.}\label{fig:num_expert}
\end{minipage}\hfill
\begin{minipage}[b]{.45\textwidth}
\includegraphics[width=\textwidth]{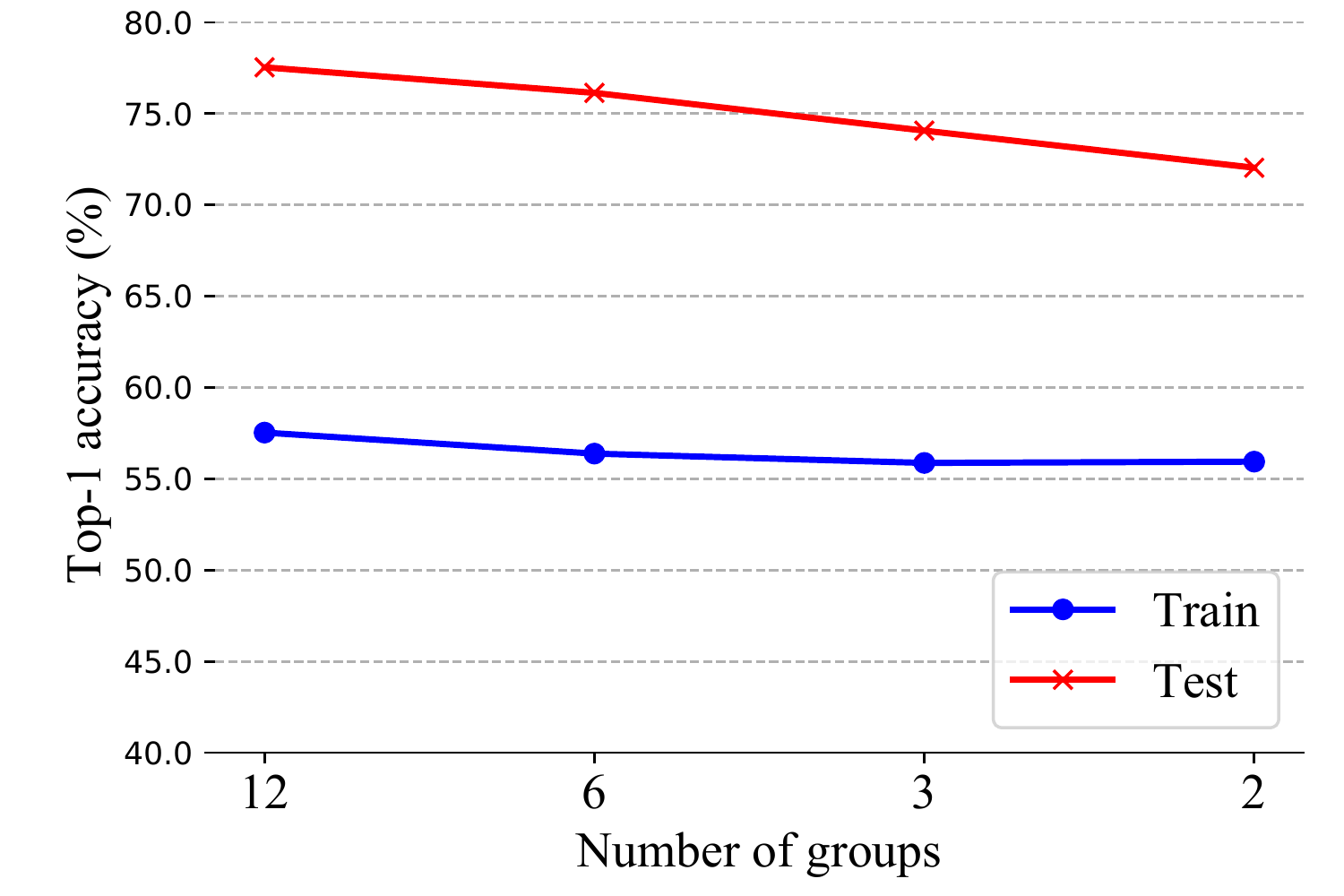}
\caption{Results of scaling the number of groups.}\label{fig:num_group}
\end{minipage}
\end{figure*}

\subsection{MoE Analysis}

To investigate the reason why MoE cannot scale well on smaller datasets like ImageNet-1K, we conduct two sets of experiments on ViT-MoE and WideNet, respectively. Given following hyper-parameters: (1) Number of training images $N_I$; (2) Number of patch tokens per image $N_p$; (3) Number of experts in each transformer block $E$; (4) Capacity ratio $C$; (5) Number of experts selected in each transformer block $K$, as we usually use a large $\lambda$, we can assume few tokens would be dropped when we are using $C$ slightly larger than $1.0$. Then, we can approximate $T \approx \frac{N_I N_p K}{E}$. Existing works~\citep{riquelme2021scaling,yang2021exploring} have shown that decreasing $N_I$, $N_p$, $K$ and $C$ can induce a performance drop. In the first set of experiments of this section, we scale the number of experts in every transformer block $E$ to control the tokens fed into each expert on ImageNet-1K.

Results are shown in Fig.~\ref{fig:num_expert}. We observe that more experts (trainable parameters) lead to overfitting although more experts mean stronger modeling capacity. Training accuracy is lower than testing accuracy because of data augmentation we introduced in the Experimental Settings Section.


To further verify that each expert requires varied tokens to train, we conduct the second set of experiments on WideNet. We define the transformer blocks using the same routing assignment that belongs to one group. To change the input diversity of each expert, each group includes more than one transformer block. That is, the hidden representations corresponding to the same token would be fed into the same expert within the same group. We set $G$ groups in total and each group includes $\frac{D}{G}$ transformer blocks, where $D$ is the number of transformer blocks.  

As shown in Fig.~\ref{fig:num_group}, when we use fewer groups, which means we have fewer routing operations, there is an obvious performance drop. We can suggest less diversified tokens are fed to each expert because fewer groups mean less routing and assignments. Therefore, more diversified tokens are required to train MoE based models on smaller datasets. More importantly, such results show the effectiveness and necessity of our design, routing at every transformer block.

\subsection{Layer Norm Analysis} 

We are to analyze the reason why individual layer normalization can improve performance in this section. Compared with the vanilla transformer structure, we share trainable matrixes in MHA and FFN (or MoE) layer across transformer blocks. The modeling capacity is compressed due to the same trainable parameters across blocks. Although WideNet uses the MoE layer to replace the FFN layer to improve capacity, different blocks are still using the same trainable parameters. Therefore, in WideNet, we encourage more diversified input to represent various semantics in different transformer blocks. Compared with vanilla ViT, we expect a larger variance of trainable vectors $\gamma$ and $\beta$ across blocks. In this section, we are interested in layer normalization before MoE or FFN.

Therefore, for $i^{\mathrm{th}}$ element of trainable vector $\gamma$ or $\beta$ in $j^{\mathrm{th}}$ block, we compute the distance between this element and all other elements of all vectors from other blocks. Taken $\gamma$ as example, we can formulate the value $y$ we are interested in as:

\begin{equation}
y= \frac{1}{MN^2} \sum_{j=1}^N  \sum_{m=1}^M \sum_{n=1}^N I_{(j\not = n)} | \gamma_{ij} - \gamma_{mn} |
\end{equation}
where $N$ is the number of transformer blocks, $M$ is the dimension of vector $\gamma$ or $\beta$.

\begin{figure*}
\centering
\begin{minipage}[b]{.45\textwidth}
\includegraphics[width=\textwidth]{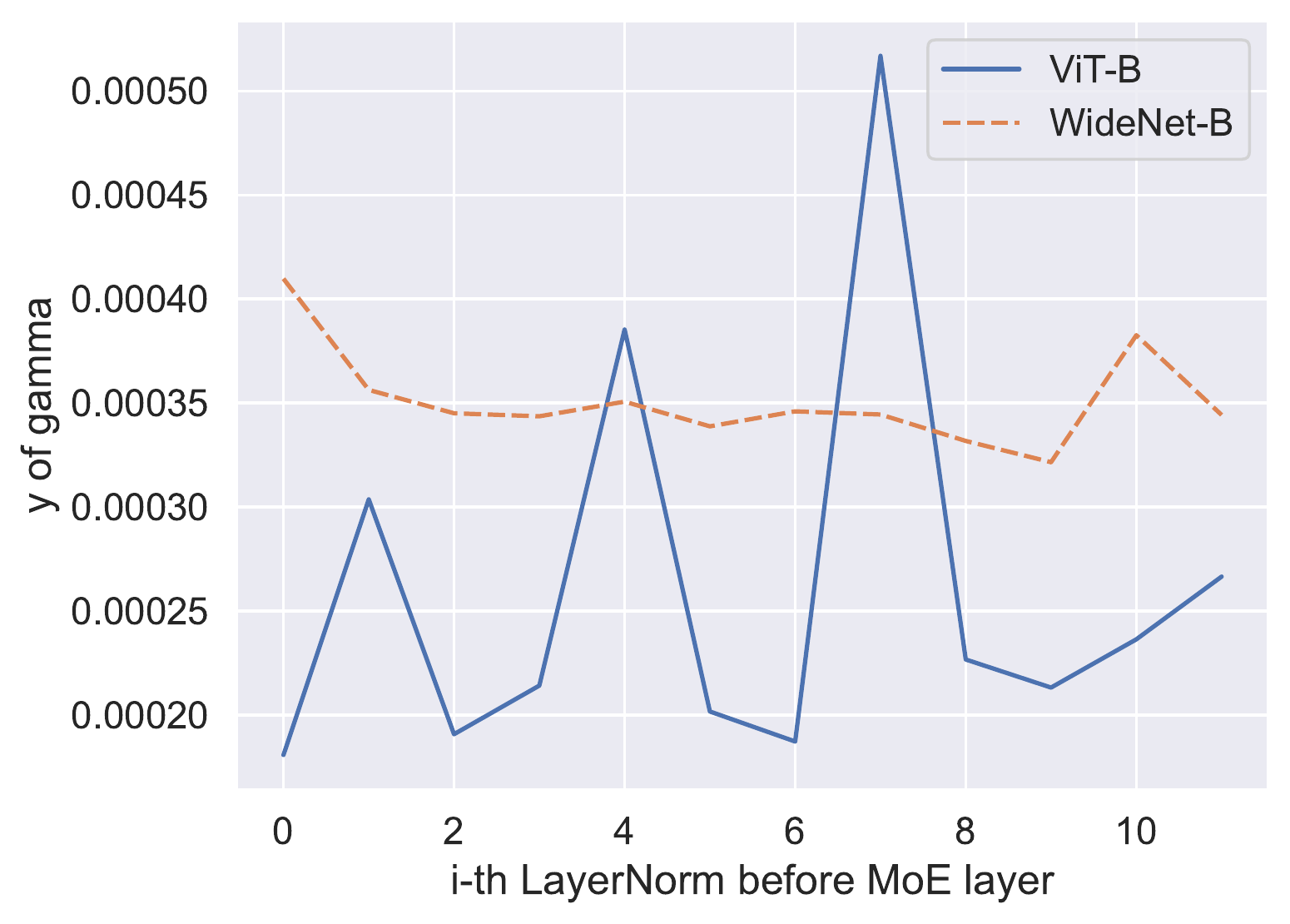}
\caption{Divergence of $\gamma$ with LayerNorm layers.}\label{fig:div_gamma}
\end{minipage}\hfill
\begin{minipage}[b]{.45\textwidth}
\includegraphics[width=\textwidth]{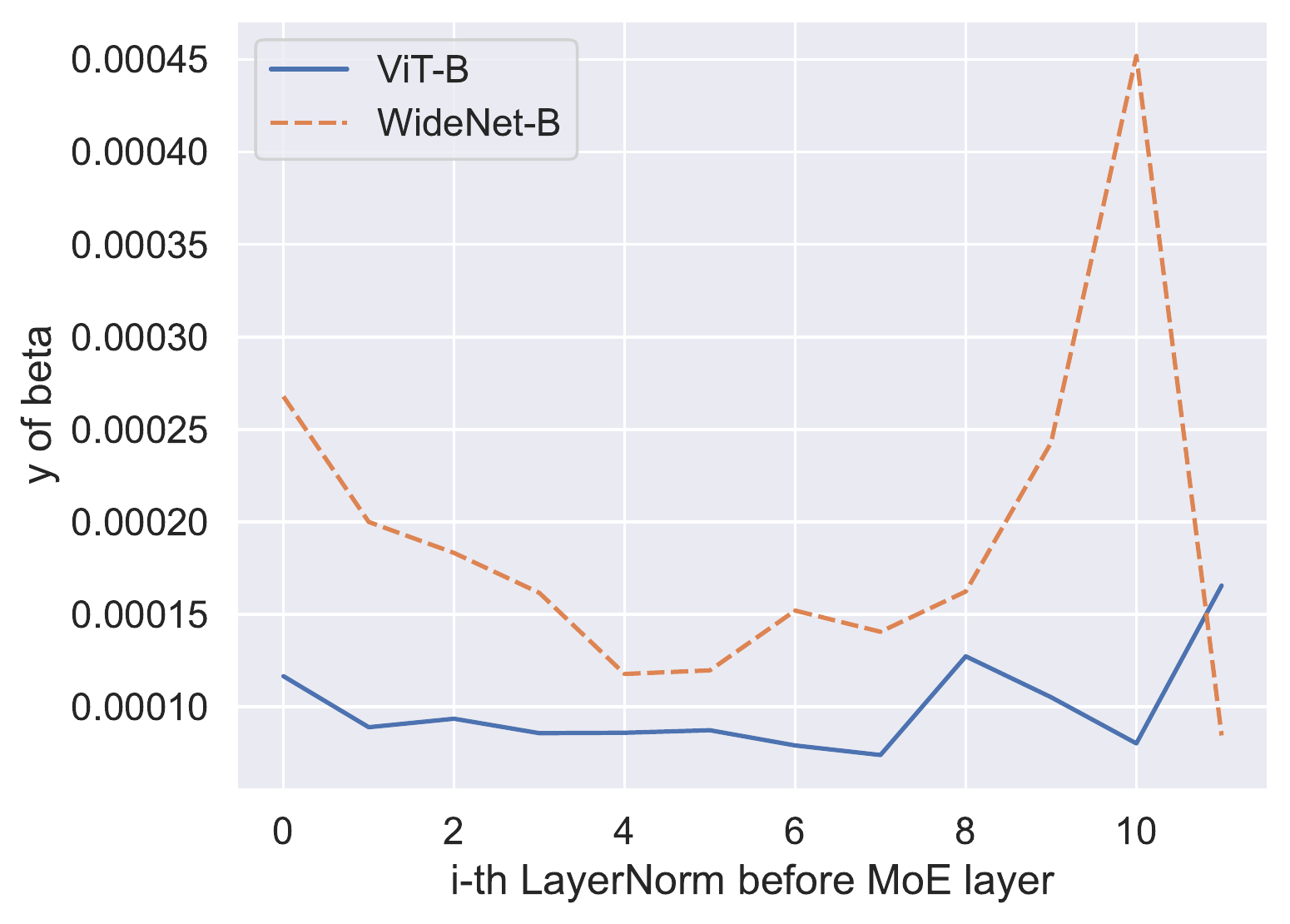}
\caption{Divergence of $\beta$ with LayerNorm layers.}\label{fig:div_beta}
\end{minipage}
\end{figure*}

In Fig.~\ref{fig:div_gamma} and Fig.~\ref{fig:div_beta}, we can observe that both $\gamma$ and $\beta$ in WideNet have larger $y$ than those in ViT, which means MoE receives more diversified input than ViT. Such result proves our assumption that individual normalization layer can help to model various semantics model with shared large trainable matrixes like MoE. 

\begin{table}[t]
\centering
\caption{Results of ablation study on ImageNet-1K to to investigate the contributions of our three key modifications (\ie Independent Layer Normalization, scaling width with MoE layer and compressing depth with parameter sharing).}
\label{tbl-abaltion-pretraining-ImageNet}
\begin{tabular}{l|ll}
\toprule
Model                                   & Top-1 & Parameters \\ \midrule
WideNet-B               & 77.5 & 29M        \\
~w/ shared Layer Norm & 76.3          & 29M                 \\
~w/o MoE layer                    & Nan            & 9M                  \\
~w/o parameter sharing            & \textbf{77.9} & 128M                \\ \midrule
WideNet-L              & \textbf{79.5} & 40M                 \\
~w/ shared Layer Norm    &      78.3          &   40M                  \\
~w/o MoE layer                    & 76.9          & 15M        \\
~w/o parameter sharing            & 77.4          & 406M                \\ \midrule
WideNet-H              & \textbf{80.1} & 63M                 \\
~w/ shared Layer Norm   &     76.6           &     63M                \\
~w/o MoE layer                    & 79.0          & 23M        \\
~w/o parameter sharing            & OOM            &      \\
\bottomrule
\end{tabular}
\end{table}

\begin{table*}[t]
\centering
\caption{Results of ablation study on ImageNet-1K to evaluate our WideNet with comparable speed or computation cost. \#Blocks is the number of transformer blocks. FNN dim means the dimension of FFN layer. Para Sharing is whether we shared parameters across transformer blocks. Time denotes to TPUv3 core days.}
\label{tbl-abaltion-2-pretraining-ImageNet}
\begin{tabular}{l|llllll}
\toprule
Model                    & \#Blocks    & FNN dim   & Para Sharing   & Top-1 & \#Para & Time \\ \midrule
ViT-L                     & 24     & 4096    & $\times$  & 77.5 & 305M       & 0.08K         \\
ViT-L                     & 24     & 4096    & $\surd$  & 76.9 & 15M       & 0.07K         \\
WideNet-L                 & 12     & 4096    & $\surd$  & \textbf{78.2} & 40M        & 0.07K          \\ \midrule
ViT-L    & 24     & 8192   & $\surd$  & 75.8 & 24M        & 0.09K         \\
WideNet-L                   & 24    & 4096   & $\surd$  & \textbf{79.5} & 40M        & 0.14K       \\
\bottomrule
\end{tabular}
\end{table*}

\subsection{Ablation Study: Contributions of key modifications}
We first conduct the ablation study to investigate the contributions of our three key modifications (\ie Independent Layer Normalization, scaling width with MoE layer, and compressing depth with parameter sharing). The results are reported in Table~\ref{tbl-abaltion-pretraining-ImageNet}.

We first replace the individual layer normalizations with the shared ones. We can observe there is a performance drop with almost the same trainable parameters. Such observation shows the effectiveness of our design. In addition, we recover the MoE layer to the FFN layer. Without the MoE layer, the training would be extremely difficult with much less trainable parameters. For example, WideNet-B without MoE layer encounters gradient explosion, and there is a significant performance drop. Finally, without parameter sharing across transformer blocks, we can also observe a slight performance drop and significant parameter increment. For WideNet-H without parameter sharing, it encounters out-of-memory when training on 256 TPUv3 cores.

\subsubsection{Ablation Study: Comparison with comparable speed or computation cost}
As we set the number of selected experts $K$ as 2 and capacity ratio $C$ as 1.2 in WideNet, there is extra computation cost than vanilla ViT. Therefore, we conduct a second set of ablation studies to evaluate our WideNet with comparable speed or computation cost with the baselines.  

As shown in Table ~\ref{tbl-abaltion-2-pretraining-ImageNet}, compared with ViT-L, WideNet-L is more computation expensive. We can observe a training time increment. However, when WideNet-L uses fewer transformer blocks (\ie 12 blocks) than ViT-L, WideNet-L outperforms ViT-L by $0.7\%$ with slightly less training time and $13.1\%$ parameters, and, similarly, there is a larger performance improvement than ViT-L with parameter sharing. We also scale ViT-L using parameter sharing to a wider FFN layer. Then, for each token, ViT-L would have comparable computation with WideNet-L setting $K$ as 2. We can see scaling to more trainable parameters and FLOPs cannot improve the performance of ViT-L, which also shows the effectiveness and necessity of our framework. Although ViT-L has a comparable computation cost with WideNet for each token, WideNet still spends more training time per epoch. According to our experiments, there are two reasons, \ie routing operation and $C > 1.0$. We leave optimize this as our future work.










\section{Conclusion}

In this paper, we propose to go wider instead of deeper for more efficient and effective parameter deployment. We implement this plug and play framework as WideNet. Especially, WideNet first compresses trainable parameters along with depth by parameter-sharing across transformer blocks. To maximize the modeling ability of each transformer block, we replace the FFN layer with the MoE layer. Then, individual layer normalization provides a more parameter-efficient way to transform semantic representations across blocks. We show that WideNet achieves the best performance by less trainable parameters on both computer vision and natural language processing backbones. In particular, on ImageNet-1K, our best model achieves $80.1$ Top-1 accuracy with only $63M$ parameters, which outperforms ViT and ViT-MoE by a large margin. On four natural language processing datasets, WideNet outperforms ALBERT by a large margin and surpass BERT with less trainable parameters. Also, the investigation shows the reason why MoE cannot scale well on smaller datasets. That is, each expert requires enough tokens to train. Moreover, we verified that individual normalization can transform hidden representations to other domains for more diversified semantics. In summary, we show that there is a great potential of this framework to train more parameter-efficient models. 

\bibliography{aaai22}

\clearpage

\appendix
\section{Appendix}
\label{appendix}

\subsection{Finetuning results on computer vision}\label{appendix:cv-finetune}

\begin{table}[ht]
\centering
\caption{Results on Cifar10 finetuning.}
\label{tbl-main-finetune-Cifar10}
\begin{tabular}{l|l l}
\toprule
Model     & Parameters   & Cifar10\\ \midrule
ViT-B               & 85M       &  98.3      \\
ViT-L              & 305M         &   98.2    \\ \midrule
ViT-MoE-B           & 126M          &  98.5   \\
ViT-MoE-L           & 404M        &   98.5  \\ \midrule
WideNet-B         & 27M        &     98.4    \\
WideNet-L          & 38M     &       \textbf{98.8}    \\
\bottomrule
\end{tabular}
\end{table}

Results in Table~\ref{tbl-main-finetune-Cifar10} shows that our WideNet can transfer better performance from pretraining to finetuning. WideNet-L, which outperforms all baselines in pretraining, is still the best model in finetuning.

\subsection{Hyper-parameters of experiments on computer vision}\label{appendix:cv-hyper}

\begin{table}[ht]
\centering
\caption{Hyper-parameters on ImageNet-1K pretraining and Cifar10 finetuning.}
\label{tbl-hyper-parameter-pretrain}
\begin{tabular}{l|l l}
\toprule
Parameter                  & ImageNet-1K  & Cifar10   \\ \midrule
Epoch                     & 300     & 100     \\
Warmup Epochs             & 30    & 0       \\
Batch Size                & 4096  & 512      \\
Learning rate             & 0.01  & 0.03  \\
Weight Decay              & 0.1   & 0  \\
Dropout                   & 0.1   & 0.1    \\ 
Label smoothing           & 0.1    & 0     \\
Mixup prob.               & 0.5   & 0.5     \\
\bottomrule
\end{tabular}
\end{table}

The value of hyper-parameters on computer vision experiments is shown in Table ~\ref{tbl-hyper-parameter-pretrain}. On ImageNet-1K cosine learning rate decay is used after 30 warmup epochs. Please note all models are using the same hyper-parameters of Table ~\ref{tbl-hyper-parameter-pretrain}, which also shows the robustness of our model. 

\subsection{Hyper-parameters of experiments on natural language processing}
\label{appendix:nlp-hyper}

\begin{table}[ht]
\centering
\caption{Hyper-parameters on downstream NLP tasks.}
\label{tbl-hyper-parameter-nlp-pretrain}
\begin{tabular}{l|l l l}
\toprule
Parameter                  & SQuAD1.1/2.0 & MNLI & SST2  \\ \midrule
Steps                     & 3649/8144 & 10000 & 5234     \\
Warmup              & 365/814   & 1000 & 314     \\
Batch Size                & 48 & 128 & 128     \\
LR             & 5e-5/3e-5 & 3e-5 & 4e-5  \\
$C$              & 2.0/1.2  & 1.2 & 1.2 \\
$\lambda$                   & 0/0.01  & 0.01 & 0.01  \\ 
Dropout           & 0.1/0   & 0  & 0   \\
Max Length               & 384/512  & 512 & 512  \\
\bottomrule
\end{tabular}
\end{table}

As shown in Table~\ref{tbl-hyper-parameter-nlp-pretrain}, we use larger capacity ratio $C$ on SQuAD1.1 as limited training data may induce unbalanced token distribution. Therefore, we set $\lambda$ as 0 and $C$ as 2.0.

\end{document}